\documentclass[10pt,twocolumn,letterpaper]{article}

\usepackage[pagenumbers]{cvpr} 

\definecolor{cvprblue}{rgb}{0.21,0.49,0.74}
\usepackage[pagebackref,breaklinks,colorlinks,allcolors=cvprblue]{hyperref}

\usepackage{booktabs} 
\usepackage{pifont}   
\usepackage{graphicx} 
\usepackage{multirow}

\newcommand{\cmark}{\ding{51}}
\newcommand{\xmark}{\ding{55}}
\usepackage{url}

\def\paperID{*****} 
\def\confName{CVPR}
\def\confYear{2026}

\title{CineLOG: A Training Free Approach for Cinematic Long Video Generation}

\author{Zahra Dehghanian, Morteza Abolghasemi, Hamid Beigy, Hamid R. Rabiee\\
Sharif University of Technology\\
Tehran, Iran\\
{\tt\small \{zahra.dehghanian97, abolghasemi, beigy, rabiee\}@sharif.edu}
}

\begin{document}
\maketitle
\begin{abstract}
Controllable video synthesis is a central challenge in computer vision, yet current models struggle with fine grained control beyond textual prompts, particularly for cinematic attributes like camera trajectory and genre. Existing datasets often suffer from severe data imbalance, noisy labels, or a significant simulation to real gap. To address this, we introduce CineLOG, a new dataset of 5,000 high quality, balanced, and uncut video clips. Each entry is annotated with a detailed scene description, explicit camera instructions based on a standard cinematic taxonomy, and genre label, ensuring balanced coverage across 17 diverse camera movements and 15 film genres. We also present our novel pipeline designed to create this dataset, which decouples the complex text to video (T2V) generation task into four easier stages with more mature technology.  To enable coherent, multi shot sequences, we introduce a novel Trajectory Guided Transition Module that generates smooth spatio-temporal interpolation. Extensive human evaluations show that our pipeline significantly outperforms SOTA end to end T2V models in adhering to specific camera and screenplay instructions, while maintaining professional visual quality. All codes and data are available in our \href{https://cine-log.pages.dev/}{Webpage}.
\end{abstract}    

\section{Introduction}


Video synthesis has become one of the most dynamic and rapidly evolving areas in computer vision and artificial intelligence \cite{ma2025controllable}. With the recent advances in generative models \cite{ma2025video}, a growing number of methods are being proposed almost daily, aiming to push the boundaries of visual fidelity, temporal coherence, and semantic alignment \cite{imagen42025, runway2025, veo32025}. Beyond simply producing visually plausible sequences, a central challenge lies in controllable video generation, where the model is guided by explicit conditions to better match user intent \cite{wang2025ati}. 


Different controllable conditions enable models to generate content that more precisely reflects user intent \cite{ma2025controllable}. While extensive research has focused on aligning video synthesis with textual prompts \cite{zhang2025cameractrl,wang2025transpixeler,wang2025lingen}, ensuring semantic consistency with the provided descriptions, other controllable parameters have received comparatively less attention. Attributes such as genre, stylistic preferences, and cinematic factors like camera trajectory or viewpoint dynamics are important in shaping the narrative quality of generated videos. However, these dimensions of control remain underexplored compared to prompt-driven conditioning.


It is also worth noting that jointly controlling multiple aspects of video generation, such as scene composition, subject appearance, genre, and camera trajectory, offers a powerful pathway toward producing content that is both realistic and richly aligned with user intent. However, most existing approaches have focused on a limited subset of these control signals, largely constrained by the availability and quality of training data. Recent efforts have explored extending controllability beyond text prompts, but they still suffer from challenges such as simple thresholding, heuristic labeling, or insufficient annotation \cite{geng2025motion,men2024mimo,yin2023dragnuwa,zhang2025recapture}. These limitations often prevent models from capturing complex motion patterns or rich cinematic variation, ultimately constraining controllability and dataset diversity \cite{dehghanian2025camera}.


To address the aforementioned gaps, we focus on constructing a dataset that unifies multiple control signals in a rich and professionally curated manner. In this work, we introduce \textbf{CineLOG}, a large scale dataset comprising 5,000 high quality, balanced, and uncut video clips. The dataset is designed to provide diversity across genres, camera alignments, and scene decoupage, ensuring broad coverage of cinematic styles and narrative structures. Each entry in \textbf{CineLOG} is annotated with a detailed scene description, explicit camera instructions, and a categorical genre label, enabling fine grained and multi dimensional controllability in video generation.

\section{Related Work}
\label{sec:related_work}

Controllable video generation depends on both the availability and the \emph{structure} of training data especially for camera trajectory control and genre/style conditioning. Prior datasets fall into three broad families: (i) synthetic datasets with full 6-DoF supervision, (ii) game engine datasets offering photorealistic rendering and rule based cinematography, and (iii) real world datasets with camera annotations (automatic or expert labeled). While each has contributed to the field, they possess inherent limitations that motivate the need for a new, comprehensively annotated resource.

\subsection{Synthetic Datasets}

Synthetic datasets, generated using 3D rendering software, offer perfect 6-DoF control over scene elements and camera parameters. Prominent examples include LensCraft \cite{dehghanian2025lenscraft}, which provides precise camera control labels through a standardized cinematography language; SynFMC \cite{shuai2025free}, which supplies full 6D pose annotations for both cameras and dynamic objects under diverse motion rules; and the CCD \cite{jiang2024cinematographic}, which focuses on generating cinematographically plausible camera trajectories in virtual scenes by combining diffusion and transformer-based modeling conditioned on high-level text descriptions and optional keyframes. Similarly, PoseTraj-10K \cite{ji2025posetraj} introduces large scale synthetic trajectories paired with 3D object poses to supervise pose aware motion control.The primary drawback of these datasets is the significant simulation-to-real gap. The generated content often lacks the photorealism, textural complexity, and unpredictable dynamics of real-world scenes. This synthetic appearance limits the ability of models trained on them to generalize to generating authentic-looking videos.

\subsection{Game engine Datasets}
To enhance realism, researchers have increasingly turned to modern video game engines. For instance, ReCamMaster \cite{bai2025recammaster} leverages Unreal Engine 5 to capture scenes with multiple, simultaneously moving cameras for studying camera relocalization and control. Similarly, the OGameData \cite{che2024gamegen} dataset, derived from high-fidelity games, offers more naturalistic rendering and complex environments than purely synthetic alternatives.

However, this approach introduces its own set of challenges. Camera movements in game engines are often mechanical and programmatic, adhering to predefined rules such as first-person or third-person perspectives. This results in camera trajectories that lack the organic and deliberate nature of human cinematography. Furthermore, the diversity of available shots and subjects is often constrained by game mechanics, leading to a dataset skewed toward single or dual actors in third-person views, while interiors, crowds, and long continuous takes remain relatively sparse.

\subsection{Real world Datests}

Real-world datasets, sourced from video clips captured in the wild, provide the highest level of visual fidelity. Datasets such as RealEstate10K popularized training on internet realestate tours, using SLAM/SfM to extract camera trajectories for long indoor walks \cite{zhou2018stereo}. More recently, CameraBench offers labeled videos organized by a camera-movement taxonomy with benchmark protocols for open-world camera control \cite{lin2025towards}. GenDoP uses large language models (LLMs), vision language models (VLMs), and geometric signals to tag camera actions and scene dynamics at scale \cite{zhang2025gendop}. Similarly, the E.T. dataset introduces text-to-trajectory generation with a focus on character-aware camera motion \cite{courant2024exceptional}.

While this real footage closes the realism gap and often contains long, continuous shots, three significant issues persist. First, label quality is a common concern, as many works rely on noisy labels from VLMs or indirect geometric methods (e.g., SLAM \cite{goel2023humans}) rather than human annotations aligned with a consistent cinematic taxonomy \cite{zhang2025gendop,courant2024exceptional}. Second, these datasets suffer from severe imbalance; the distribution of camera movement, genres, and subject number is heavily skewed, which biases controllers toward a few common movement types and prevents generalization. Third, availability is a major barrier, as these datasets are often not publicly available or are limited in scale \cite{lin2025towards}. We detail other influential datasets with camera supervision in Appendix \ref{app: other related work}; while they contain camera annotations, they lack the deliberate cinematic camera movements essential for our task. 
To compare these datasets along with our ViDivese, we summarize the characteristics of these datasets in Table~\ref{tab:dataset_comparison}.

\begin{table*}[h!]
\centering
\caption{Comparison of Datasets for Camera Movement Analysis}
\label{tab:dataset_comparison}
\resizebox{0.7\textwidth}{!}{%
\begin{tabular}{@{}c l ccc ccc@{}}
\toprule
& \textbf{Dataset} & \textbf{\begin{tabular}[c]{@{}c@{}}Diversity\\ Subject\end{tabular}} & \textbf{\begin{tabular}[c]{@{}c@{}}Diversity\\ Scene\end{tabular}} & \textbf{Balance} & 
\textbf{\begin{tabular}[c]{@{}c@{}}Cinematic\\Caption\end{tabular}}  &
\textbf{\begin{tabular}[c]{@{}c@{}}Scene\\Description\end{tabular}}  &
\textbf{\begin{tabular}[c]{@{}c@{}}Genre\\Annot.\end{tabular}} \\

\midrule
\multirow{6}{*}{\rotatebox[origin=c]{90}{\textbf{\small{Unreal}}}} 

& SynFMC \cite{shuai2025free}              & Wide & \cmark & \cmark & \xmark & \xmark & \xmark \\
& CCD \cite{jiang2024cinematographic}       & Only Human & \xmark & \cmark & \cmark & \xmark & \xmark \\
& LensCraft \cite{dehghanian2025lenscraft} & Wide & \xmark & \cmark & \cmark & \xmark & \xmark \\
& PoseTraj-10K \cite{ji2025posetraj}        & Wide & \xmark & \xmark & \xmark & \xmark & \xmark \\
& ReCamMaster \cite{bai2025recammaster}     & Only Human & \cmark & \cmark & \xmark & \xmark & \xmark \\
& OGameData \cite{che2024gamegen}           & Only Human & \xmark & \xmark & \cmark & \xmark & \xmark \\
\midrule
\multirow{4}{*}{\rotatebox[origin=c]{90}{\textbf{\small{Real}}}} 
& RealEstate10K \cite{zhou2018stereo}  & Only Indoor & \cmark & \xmark & \xmark & \xmark & \xmark \\
& CameraBench \cite{lin2025towards}   & Wide & \cmark & \xmark & \cmark & \xmark & \xmark \\
& DataDoP \cite{zhang2025gendop}& Wide & \cmark & \xmark & \cmark & \cmark & \xmark \\
& E.T. \cite{courant2024exceptional}   & Only Human & \xmark & \xmark & \cmark & \xmark & \xmark \\
\midrule
& \textbf{CineLOG (Ours)}  & \textbf{Wide} & \textbf{\cmark} & \textbf{\cmark} & \textbf{\cmark} & \textbf{\cmark} & \textbf{\cmark} \\
\bottomrule
\end{tabular}
}
\end{table*}

Existing datasets present a fundamental trade off, forcing a choice between the precise but artificial supervision of synthetic data and the realistic but poorly labeled and imbalanced nature of real world corpora. Our dataset is designed to bridge this divide by providing: (i) balanced, taxonomy aligned coverage of all camera primitives; (ii) long, uncut, public, and diverse high fidelity scenes across various genres and subject counts, without licensing restrictions; and (iii) paired, cinematography oriented annotations detailing subjects, genres, camera movements, and initial/final views. Together, these features enable robust camera control for open world Text to Video (T2V) generation, advancing beyond prior work.

\section {CineLOG}
\label{sec: methodology}


Two long standing blockers in dataset construction diversity of content and real world visual fidelity are now largely mitigated by steady advances in T2V diffusion models and high capacity video backbones \cite{ma2025video,wang2025lingen}. Building on this,our starting point for generating dataset is pragmatic: Using rapid advances in modern image/video generative models to synthesize large volumes of controlled videos. Concretely, we first evaluated several state of the art (SOTA) T2V systems and motion aware controllers as dataset engines. In practice, current T2V solutions split into two classes: (i) open-source/commodity models \cite{wan2025wan, peng2025open} produce visually strong, temporally coherent clips but show a recurring failure mode: weak grounding to explicit cinematographic instructions, and (ii) premium/closed models \cite{KlingAI, OpenAI2024Sora} that typically yield higher perceptual quality and somewhat a bit better camera semantics, but whose API pricing and throughput limits make them impractical for generating a high scale, fully supervised dataset. Empirically, we observe that both classes still prioritize aesthetic plausibility over strict adherence to explicit cinematography directives, especially under long or multi step trajectories, lensing, or coupled pans/tilts. Qualitative samples (including prompts, camera specifications, and output shots) for both model classes are provided in Supplementary \ref{app: failure models}.

Building on this observation, we pivot from end to end T2V to a new strategy: modern T2I diffusion models exhibit stronger inductive biases for composition, shot framing, and camera instructions and follow spatial constraints more faithfully \cite{GoogleGemini2.5FlashImage2025}. This aligns with the industry practice of key-framing: a director specifies a sparse set of key frames and the cinematographer executes fluid motion between them. We operationalize the same idea by using T2I to first synthesize a storyboard \cite{jiang2024cinematographic}.

We therefore leave end to end T2V for camera controllability and factor the problem into two cooperating modules. We use a strong T2I model \cite{GoogleGemini2.5FlashImage2025} to synthesize a small set of keyframes that obey explicit cinematographic instructions for composition and framing. This exploits the superior spatial grounding of T2I systems under conditioning and structural guidance, and then pass two by two to a first last frame to video (FLF2V) generator, which predicts temporally coherent in betweens and respects the target camera path. 

\begin{figure*}
    \centering
    \includegraphics[width=\textwidth]{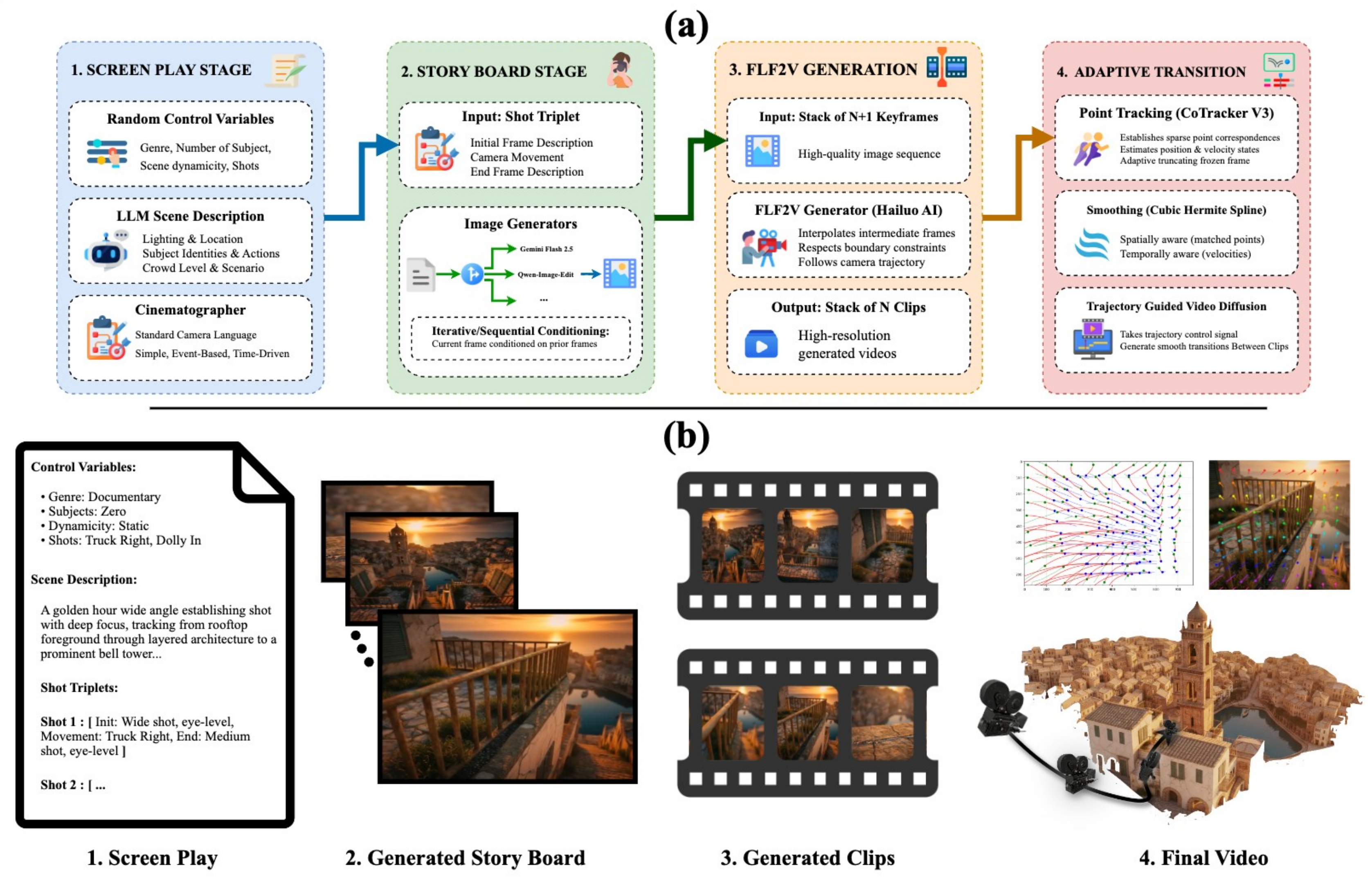}
    \caption{The CineLOG generation architecture, which decouples complex video synthesis into four stages for precise cinematic control. (a) The pipeline flow: (1) The Screen Play stage uses LLMs to generate a detailed shot plan. (2) The Story Board stage synthesizes $N+1$ keyframes. (3) FLF2V Generation creates $N$ independent clips between the keyframes. (4) The Adaptive Transition module analyzes motion and smoothly stitches the clips into a final, seamless video. (b) A sample output from each stage.}
    \label{fig: pipeline generation}
\end{figure*}

To preserve continuity across shots, we add a stage that concatenates adjacent clips and resolves spatial and temporal discontinuities. Figure \ref{fig: pipeline generation} illustrates the full pipeline from screenplay to transition tuning to generate our high quality videos for our CineLOG dataset. In the following subsections, we will dive into details of each stage in our proposed pipeline.

\subsection{Screen Play}
\label{sec: screen play}




The goal of this stage is to synthesize a compact, cinematography aware screenplay that conditions all down stream modules. A key requirement for our dataset is achieving diversity across several critical control parameters, including genre, shot count, per shot camera movements, main subject count, and scene dynamicity. We empirically observed that tasking an LLM to generate random values for these parameters resulted in significant mode collapse, with the model repeatedly defaulting to common, high frequency choices. So to ensure a balanced distribution and avoid this sampling bias, we instead explicitly pre select the values for these discrete control variables under a uniform balance constraint. This set of parameters is then passed as direct conditional input to the LLM, guaranteeing full coverage of our desired cinematic envelope. The complete taxonomy and range of options for each control signal are detailed in Supplementary \ref{app: screenplay details}.

Given these control signals, a LLM produces a structured scene record. The LLM is prompted to return a structured detailed record that needs for generating a precise view of the scene includes lighting description, location description, subject list with identities and visual attributes, subject actions with fine grained verbs, subject location in the scene, crowd level with approximate counts or density, and a scenario paragraph that spans the shots. This representation encourages stable composition and reduces prompt drift under later conditioning.


A second LLM, acting as the cinematographer, is then tasked with translating the abstract camera movement instructions into a format suitable for the Storyboard generation. We observed that the T2I model fails to accurately synthesize the final keyframe when given insufficient visual context; simply providing the cinematic instruction (e.g., dolly in) or its high level description is inadequate for the model to infer the precise resulting composition. The cinematographer's role is therefore to interpret the specified camera movement and generate a new, detailed textual description of what the camera will see at the end of that motion. For example, dolly in is translated into a description like, A tight close up on the character's face, showing their eyes in detail, with the background now out of focus. This translated description provides a much stronger conditional signal to the T2I model, enabling it to generate the final shot's keyframe with high precision.

The output of the screenplay stage is a sequential list of per shot triplets, structured as [Shot Init, Camera Movement, Shot End]. For the first shot, the Shot Init is the detailed scene record generated by the first LLM, and the Shot End is the translated visual description from the cinematographer LLM based on the shot's Camera Movement. For any subsequent shots, this process is chained to ensure continuity: the Shot Init for the current shot is precisely the Shot End description from the previous shot. The cinematographer then generates a new Shot End for the current shot based on this new starting description and its corresponding Camera Movement. This triplet structure provides clear, descriptive boundaries for the storyboard generation stage.

\subsection{Story Board}




The goal of this stage is to operationalize the industry practice of keyframing \cite{hart2013art}, where a director specifies a sparse set of keyframes to guide fluid motion execution. This module synthesizes the high fidelity keyframes that serve as the precise visual anchors for the subsequent video generation stage. The input for this module is the per shot triplet list provided by the previous Screenplay stage.

The synthesis process begins with the first shot's Shot Init description. This initial step is a pure T2I task, for which we utilize a SOTA T2I generator which is conditioned on the full scene narration and the detailed Shot Init prompt to generate the first keyframe with high fidelity.

For all subsequent keyframes, the task transitions from T2I to conditioned Image to Image (I2I) generation. To ensure visual and semantic continuity, each new keyframe must be generated by conditioning on both the previous keyframe and the new textual Shot End description. This sequential conditioning is critical for maintaining subject identity and appearance across shots and reducing compositional drift.

During our analysis, we made a critical observation: no single SOTA I2I model excels at all 17 cinematic camera movements defined in our taxonomy. We found significant performance variance across models.

Given that no single model could serve as a universal generator, we implemented a dynamic routing strategy to ensure the highest possible quality for every shot. We first benchmarked all SOTA I2I models by tasking them with generating 50 use cases for each distinct camera movement. The results of this evaluation are presented in Table \ref{tab:camera_comparison}. Values represent the average score (out of 10) from 10 qualitative users, based on three criteria, (1) preserving scene, (2) following camera instruction, and (3) following story narration. For dynamic camera, we choose random direction eg, choose pedestal up between pedestal up or down.


Based on this analysis that there were no best model that win all, so we decide to dynamically route the I2I generation for each shot to the model that achieved the best performance for that specific camera movement. This ensures that every keyframe is synthesized by the most suitable generator, creating an influential and high fidelity storyboard.

The final output of this stage is a stack of $N+1$ high quality images, where $N$ is the shot count. This sequence of keyframes represents the complete visual storyboard, which is then passed to the FLF2V Generation module.



\subsection{FLF2V Generation}



The FLF2V generator's core function is to predict the smooth, intermediate video frames that respect both the compositional constraints of the boundary keyframes and the explicitly defined prompt \cite{wan2025wan}. This architecture factors the complex T2V problem into two cooperating modules, leveraging the superior spatial grounding of text to image (T2I) systems for keyframe synthesis and delegating the temporal coherence task to the FLF2V module.


In this stage we use a SOTA FLF2V generation module to achieve temporal coherence and precise camera control. Specifically, the Hailuo AI model \cite{HailuoAl2025}) is tasked with interpolating the intermediate video frames. The input to this module consists of two distinct components generated in previous stages: (1) the sequence of keyframes, generated by the Story Board stage, and (2) the corresponding camera trajectory triplet, ⟨Shot Init,Camera Movement,Shot End⟩, provided by the Screen Play stage. 
This process effectively converts the static visual constraints and the explicit motion instructions into a temporally continuous video segment, and the generated sequence not only maintains visual fidelity but also adheres

The output of this stage is a list of high resolution video clips that satisfies the initial and terminal spatial constraints, to ensure maximum visual smoothness and to mitigate any potential frame level artifacts arising from the generative process, we employ a final post processing for seamlessly long video generation.

\subsection{Adaptive Seamless Transition}
\label{sec: transition}

The generation of long, contiguous video sequences often relies on stitching together multiple shorter clips. While generating each clip independently is feasible, a critical challenge arises at the cut boundaries. A simple concatenation of two videos, even when sharing a middle frame or a short overlapping segment, results in a noticeable discontinuity in the perceived camera movement. This visual artifact, often referred to as a "jump cut" or "hiccup", severely compromises the desired smoothness and naturalness of the resulting film.

The root of this problem stems from two interconnected cinematic challenges inherent in video generation models:
\begin{itemize}
\item Inconsistent Velocity: Video generators tend to accelerate the camera trajectory at the starting frames and decelerate toward the final frames. When two such clips are appended, this deceleration acceleration cycle creates the impression that the camera briefly stops or freezes at the transition timestamp.

\item Divergent Trajectories: Crucially, the direction and acceleration of the camera's motion are different between merged clips. This sudden change in directional momentum will result in a jarring visual jolt and , destroy the illusion of continuous cinematic motion..

\end{itemize}

A naive approach to mitigate these issue might involve cropping the slow, end and start frames from the two adjacent clips and using blindly standard video frame interpolation or FLF2V models to bridge the gap. However, this solution is fundamentally flawed: this process creates a third, short film that is then spliced between the original two. As the interpolated segment's trajectory is not explicitly connected to the underlying motion dynamics of the neighboring clips, the problem of discontinuity simply shifts, now existing at the two new connection points.

It should be mentioned that there exists a huge volume of work in Video Frame Interpolation, but most of these methods approach the task as an FLF2V problem with a critical limitation for our use case \cite{wu2024perception, guo2024generalizable}.
Their primary focus during training is on interpolating subject and character movement \cite{kye2025acevfi}, often with an implicit assumption that the underlying camera motion is continuous and consistent. Consequently, these models have challenges in scenarios where the camera motion exhibits a massive directional or velocity change \cite{kye2025acevfi}, like ours.

To truly solve this challenge, we need a model that not only performs frame completion but also understands and respects the full underlying camera trajectory of both adjacent video segments. This model must be capable of generating a transition segment whose camera motion smoothly and naturally bridges the gap between the two distinct trajectories. 



To handle this critical challenge and ensure the cinematic integrity of our long form video generation, we propose a Trajectory Guided Transition Module. Near to simple frame generation, there exist several works that takes control signals to guide their generation process \cite{briedis2025controllable, wang2025framer}. Here, we use Framer model \cite{wang2025framer}  which takes as input a predefined trajectory as a condition to generate the intermediate video frames. 

However, the Framer model \cite{wang2025framer} requires a detailed trajectory control signal to ensure a smooth camera transition. To generate this signal, we first establish sparse point correspondences. Our pipeline uses the shared keyframe, where the first clip ends and the second begins, as an anchor. From this frame, we employ CoTrackerV3 \cite{karaev2025cotracker3} to track points bidirectionally: backward into the reversed first clip and forward into the normal second clip. This establishes continuous point trajectories across the two segments. 

Next, to resolve the freezing problem, we adaptively truncate the near-static frames. We analyze a window (max 30 frames) at the stitch point, defining a frame as frozen if its tracked point motion, relative to the anchor frame, is below a minimum threshold. Iterating from the clip boundaries, we truncate all frames until the first frame with significant motion is detected. The final sparse point correspondences are then computed from these established trajectories at the new, truncated boundary frames.

For each matched point, we estimate its initial state (position $p_0$, velocity $v_0$) from the first clip and its terminal state (position $p_1$, velocity $v_1$) from the second clip. We then construct a smooth, $C^1$ continuous spatio-temporal path, $P(t)$ for $t \in [0, 1]$, using Cubic Hermite Spline \cite{de1978practical}. This ensures the trajectory is spatially aware of the matched points and temporally aware of their motion in both clips. The interpolation, expanded into its standard polynomial form, is defined as:
\begin{equation}
\begin{split}
P(t) = &(2p_0 + v_0 - 2p_1 + v_1)t^3 \\
& + (-3p_0 - 2v_0 + 3p_1 - v_1)t^2 \\
& + v_0t + p_0
\end{split}
\end{equation}
The resulting dense field of interpolated point trajectories serves as the control signal. This signal is then fed to the Framer model to synthesize a visually and temporally continuous transition. 

\section {Experiments}

\label{sec:experiments}

In this section, we conduct a series of quantitative and qualitative experiments to validate the key contributions of our work. Our evaluation focuses on the fidelity and correctness of each stage in our proposed generation pipeline, from screenplay generation to seamless video transition.

\begin{figure*}[h]
    \centering
    \includegraphics[width=\textwidth]{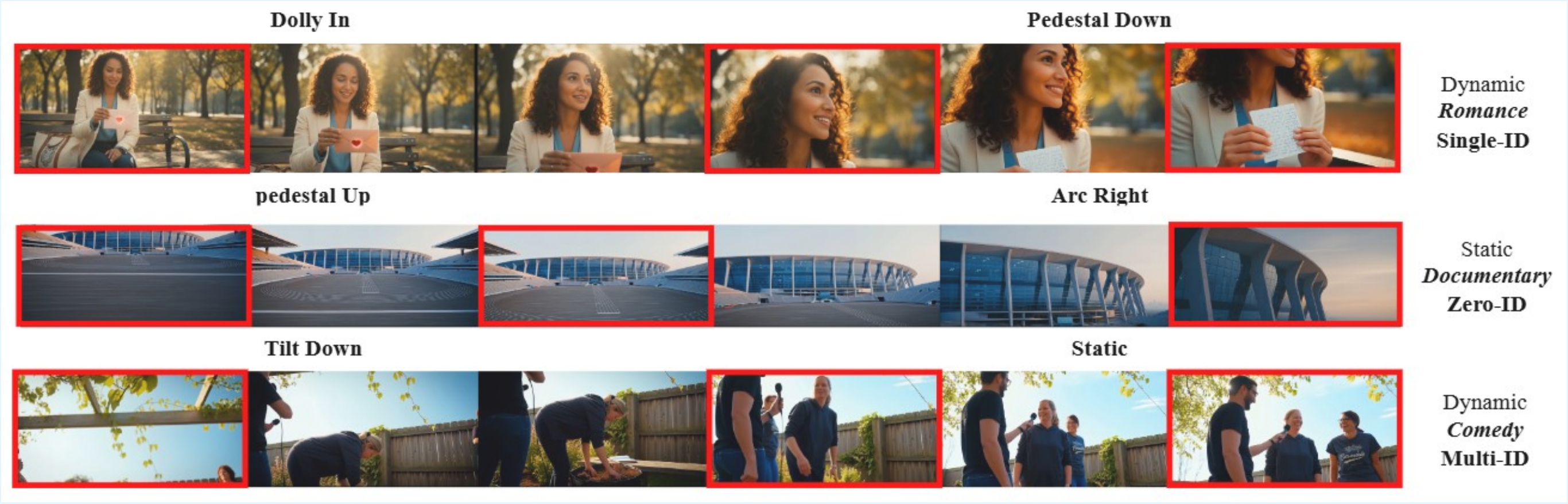}
    \caption{Visual overview of the CineLOG dataset. Keyframes are shown by red outlines. The dataset provides rich annotations, including different camera movement, genre, number of subjects, and more.}
    \label{fig: qualitative samples}
\end{figure*}

In Figure \ref{fig: qualitative samples}, we present some samples from our CineLOG dataset. Each sample includes all control signals and the explicit camera instructions for each shot, full screen play , and a series of keyframes from the final generated video. 




\subsection{Screenplay Stage Evaluation}

Due to the sequential nature of our designed pipeline, the fidelity of the entire pipeline depends on the quality of the initial screenplay, which means that we need an LLM to capably function as both a storyteller and a cinematographer. This model must correctly translate abstract control variables into structured, coherent scene descriptions and camera plans. However, technical reports and benchmarks present conflicting claims on which LLM excels at creative narrative tasks \cite{wu2025writingbench, fein2025litbench}. 

Given this ambiguity, we conducted a comparative study to determine the most suitable model for our specific task. We fed all three models the same set of 100 combinations of control signal as described in Section \ref{sec: screen play}. To mitigate evaluation bias, where an LLM might favorably judge its own output, we performed a majority voting evaluation. The structured output from each model was evaluated by all models. We measured the average binary classification accuracy for retrieving the initial control parameters and the variance of these classifications, where lower variance indicates higher inter model agreement and confidence. The result of this comparison is shown in Table \ref{tab:llm_ablation}

\begin{table*}[h]
\centering
\caption{LLM ablation for Screenplay generation. We report the parameter retrieval accuracy (\%) $\pm$ variance (Var) for each control signal, evaluated by all three models.}
\label{tab:llm_ablation}
\resizebox{0.8\textwidth}{!}{%
\begin{tabular}{l|cccc|c}
\toprule
\textbf{Model} & \textbf{Genre} & \textbf{Subject Count} & \textbf{Dynamicity} & \textbf{Shot Count} & \textbf{Average} \\
\midrule
GPT-5 mini \cite{OpenAI2025GPT5}& \textbf{94.7} $\pm$ 0.00 & \textbf{59.6} $\pm$ 0.02 & \textbf{96.5} $\pm$ 0.02 & \textbf{100.0 }$\pm$ 0.00 & \textbf{87.7} \\
Qwen 3 \cite{yang2025qwen3}& 86.0 $\pm$ 0.01 & 40.4 $\pm$ 0.02 & 87.7 $\pm$ 0.06 & 15.8 $\pm$ 0.04 & 57.5 \\
Gemini 2.5 flash \cite{Gemini2025Flash}& 89.5 $\pm$ 0.00 & 45.6 $\pm$ 0.02 & 84.2 $\pm$ 0.06 & 21.1 $\pm$ 0.02 & 60.1 \\
\bottomrule
\end{tabular}
}
\end{table*}

As shown in Table \ref{tab:llm_ablation}, GPT-5 mini achieves the highest accuracy and lowest variance, confirming its superior capability for this task. Based on this experiment we choose this model as strory teller for our screen play generation.

\subsection{Storyboard Stage Evaluation}
\label{sec:exp_storyboard}
Our pipeline operationalizes key framing using a I2I model to synthesize storyboards. Due to rapid progress in image generation models, conventional automatic metrics are no longer capable of indicating semantic correctness, controllability, or adherence to user intent. So, to ensure the quality of the outputs, from this stage onward, we use human evaluation, the most accurate and trustworthy method for comparisons.

To select the strongest model, we conducted a comparative study including \textbf{Gemini 2.5 Flash Image} \cite{GoogleGemini2.5FlashImage2025}, \textbf{Qwen Image Edit} \cite{wu2025qwen}, \textbf{FLUX Kontext Pro} \cite{batifol2025flux}, \textbf{Seedream 4} \cite{seedream2025seedream}, \textbf{Bria FIBO} \cite{gutflaish2025generating}, \textbf{SeedEdit 3.0} \cite{wang2025seededit}, and \textbf{Qwen Lora Camera} \cite{dx8152_QwenEditLoRA_2025} which is a LoRA fine tuning of ImageEdit model on some camera Instructions. The evaluation must focus on metrics needed for generating coherent, multi shot cinematic narratives. The detailed description of metrics for this stage is provided in the supplementary \ref{sup: evaluation protocol} and the results of it is shown in Table \ref{tab:camera_comparison}.

\begin{table*}[t]
  \centering
  \caption{Comparative analysis of I2I models for the Storyboard Generation stage.}
  \label{tab:camera_comparison}
  \resizebox{\textwidth}{!}{
  \begin{tabular}{l *{11}{c}}
    \toprule
    \multirow{2}{*}{\textbf{Model}} & \multicolumn{9}{c}{\textbf{Camera Adherence}} & \multirow{2}{*}{\textbf{\shortstack{Scene \\ Preservation}}} & \multirow{2}{*}{\textbf{\shortstack{Narration \\ Adherence}}} \\
    \cmidrule(lr){2-10} 
    & \textbf{Static} & \textbf{Pan} & \textbf{Tilt} & \textbf{Dolly} & \textbf{Truck} & \textbf{Pedestal} & \textbf{Zoom} & \textbf{Crane} & \textbf{Arc} & & \\ 
    \midrule
    Gemini Flash 2.5 \cite{GoogleGemini2.5FlashImage2025}    & \textbf{8.2} $\pm$ 2.2 & 4.4 $\pm$ 2.2 & \textbf{6.2} $\pm$ 2.8 & 7.7 $\pm$ 2.2 & 4.2 $\pm$ 3.6 & \textbf{7.0} $\pm$ 1.1 & \textbf{7.8} $\pm$ 2.2 & 7.2 $\pm$ 3.4 & 3.6 $\pm$ 2.0 & \textbf{8.6} $\pm$ 2.0 & 7.3 $\pm$ 3.0 \\
    Seedream 4      \cite{seedream2025seedream}       & 8.0 $\pm$ 2.6 & 4.2 $\pm$ 3.0 & 4.0 $\pm$ 2.6 & 5.3 $\pm$ 3.4 & 4.3 $\pm$ 2.1 & 7.0 $\pm$ 2.4 & 3.4 $\pm$ 2.4 & 7.0 $\pm$ 3.0 & 3.4 $\pm$ 2.0 & 8.2 $\pm$ 2.2 & 6.1 $\pm$ 3.0 \\
    Flux Kontext Pro  \cite{batifol2025flux}     & 7.4 $\pm$ 2.6 & 4.0 $\pm$ 2.6 & 3.6 $\pm$ 2.2 & 5.2 $\pm$ 3.2 & 2.6 $\pm$ 1.2 & 5.2 $\pm$ 2.6 & 4.0 $\pm$ 3.0 & 4.4 $\pm$ 2.6 & 4.6 $\pm$ 2.6 & 8.2 $\pm$ 2.0 & 5.4 $\pm$ 2.8 \\
    Bria FIBO   \cite{gutflaish2025generating}      & 5.4 $\pm$ 2.6 & 4.0 $\pm$ 2.2 & 4.4 $\pm$ 3.8 & 6.6 $\pm$ 3.2 & 5.4 $\pm$ 3.2 & \textbf{7.0} $\pm$ 2.6 & 3.6 $\pm$ 2.6 & 5.2 $\pm$ 2.4 & 3.6 $\pm$ 2.0 & 4.4 $\pm$ 2.4 & 5.5 $\pm$ 2.8 \\
    Qwen ImageEdit  \cite{wu2025qwen}     & 6.8 $\pm$ 2.6 & 5.0 $\pm$ 3.2 & 2.6 $\pm$ 1.6 & \textbf{8.2} $\pm$ 2.6 & 4.6 $\pm$ 2.8 & 5.4 $\pm$ 2.2 & 4.0 $\pm$ 3.0 & 6.4 $\pm$ 3.2 & 3.8 $\pm$ 2.4 & 7.2 $\pm$ 2.6 & 5.8 $\pm$ 2.8 \\
    SeedEdit 3.0    \cite{wang2025seededit}     & 6.6 $\pm$ 3.0 & 4.0 $\pm$ 2.8 & 4.2 $\pm$ 2.8 & 6.0 $\pm$ 3.0 & 2.6 $\pm$ 1.2 & 6.8 $\pm$ 3.2 & 5.2 $\pm$ 3.0 & 3.8 $\pm$ 1.8 & 3.6 $\pm$ 2.0 & 6.6 $\pm$ 2.8 & 5.0 $\pm$ 2.8 \\
    Qwen Lora Camera  \cite{dx8152_QwenEditLoRA_2025}    & 4.8 $\pm$ 2.8 & \textbf{7.0} $\pm$ 3.0 & 5.2 $\pm$ 3.0 & 5.8 $\pm$ 2.6 & \textbf{7.9} $\pm$ 2.8 & 6.4 $\pm$ 2.4 & 5.0 $\pm$ 3.0 & \textbf{8.4} $\pm$ 2.4 & \textbf{9.0} $\pm$ 1.6 & 7.2 $\pm$ 2.6 & \textbf{7.6} $\pm$ 2.8 \\
    \bottomrule
  \end{tabular}
  } 
\end{table*}

Our comparative analysis, detailed in Table \ref{tab:camera_comparison}, reveals a significant performance variance across models, confirming our observation that no single generator excels at all 17 cinematic movements. While Gemini Flash 2.5 demonstrates superior generalist capabilities, achieving the highest scores for Scene Preservation (8.6), the specialist Qwen Lora Camera model yields the best results for complex dynamic motions such as Pan (7.0), Truck (7.9), and Crane (8.4). Given this finding, we implemented a dynamic routing strategy, leveraging Gemini Flash 2.5 for its strong contextual consistency while routing specific dynamic movement instructions to the Qwen Lora Camera to ensure the highest possible fidelity for every generated keyframe.

\subsection{FLF2V Generation Evaluation}
\label{sec:exp_flf2v}
As established in our pipeline design (Section \ref{sec: methodology}), current end to end T2V models struggle to faithfully execute complex, multi shot cinematographic instructions, often prioritizing visual plausibility over strict prompt adherence \cite{veo32025, HailuoAl2025}. To validate the superiority of our proposed T2I$\rightarrow$FLF2V pipeline, we conduct a rigorous human evaluation. We compare our final video shots against two SOTA T2V models: \textbf{Hailuo 2.3} \cite{HailuoAl2025} and \textbf{Veo 3.2} \cite{veo32025}. We provided the full screenplay description (including scene and camera instructions) from 100 random samples directly to these T2V models.
The detailed description of the evaluation protocol for this stage is provided in the supplementary \ref{sup: evaluation protocol}.

\begin{table}[h]
\centering
\caption{Human evaluation comparing our T2I$\rightarrow$FLF2V pipeline against end to end T2V models. We report mean user scores (0-10) and binary correctness accuracy (\%).}
\label{tab:video_gen_comparison}
\resizebox{0.95\columnwidth}{!}{%
\begin{tabular}{l|ccc}
\toprule
\textbf{Evaluation Metric} & \textbf{CineLOG} & \textbf{Hailuo 2.3} & \textbf{Veo 3.2} \\
\midrule
Screenplay Adherence & \textbf{7.6} $\pm$ 1.1 & 6.2 $\pm$ 1.2 & 5.6 $\pm$ 1.1 \\
Camera Adherence & \textbf{8.3} $\pm$ 1.1 & 5.5 $\pm$ 1.2 & 3.8 $\pm$ 1.0 \\
Realism Score & \textbf{7.2} $\pm$ 1.2 & \textbf{7.2} $\pm$ 1.3 & 6.4 $\pm$ 1.3 \\
Movement Naturalness & \textbf{7.7} $\pm$ 1.2 & 7.4 $\pm$ 1.2 & 6.5 $\pm$ 1.4 \\
Geometric Stability & \textbf{8.1} $\pm$ 1.1 & 7.4 $\pm$ 1.3 & 6.8 $\pm$ 1.4 \\
\midrule
Genre Correctness & \textbf{93.5\%} & 62.1\% & 78.3\% \\
Subject Count Correctness & \textbf{90.8\%} & 65.4\% & 68.2\% \\
Dynamicity Correctness & \textbf{89.3\%} & 51.0\% & 54.5\% \\
Shot Count Correctness & \textbf{91.2\%} & 36.2\% & 41.3\% \\
\bottomrule
\end{tabular}
}
\end{table}

As shown in Table \ref{tab:video_gen_comparison}, our pipeline significantly outperforms end to end T2V models, especially in the critical areas of camera and screenplay adherence. While the T2V models achieve comparable Realism Score, they fail to correctly implement the specific camera motion instructions and shot count, validating our pipeline's design. Also it should be mentioned that the average long of our video is 12.02 in comparison to 8 (Veo 3.2) and 6 (hailou AI).

\subsection{Adaptive Seamless Transition Evaluation}
\label{sec:exp_transition}

Validating the effectiveness of our Trajectory Guided Transition Module (Section \ref{sec: transition}) is crucial. We conducted a comparative human study to prove its superiority over naive baselines. The details of the evaluation metrics is provided in supplementary \ref{sup: evaluation protocol}. Evaluators were shown 50 multi shot sequences generated using three different transition methods:
\begin{itemize}
    \item \textbf{Simple Concatenation:} Attaching two video clips directly.
    \item \textbf{Blind Interpolator:} A SOTA video frame interpolation model Framer \cite{wang2025framer} (without any trajectory guidance).
    \item \textbf{w.o. adaptive windowing:} Truncate fixed 20 frame, from both clips and then generate the control signal. 
    \item \textbf{Trajectory Aware (Our):} Our proposed method using Cubic Hermite interpolation to guide the Framer model.
\end{itemize}
We asked for ratings (0-10) on three criteria: \textbf{Geometric Stability}, \textbf{Smoothness}, and \textbf{Visual Pleasingness}. We also performed a head to head ranking, asking evaluators to sort the videos from best to worst. Table \ref{tab:transition} reports the mean scores of the criteria and the Win Rate is the percentage of time a method was ranked 1st. 


\begin{table}[h]
\centering
\caption{Human evaluation of the adaptive Transition module. Our trajectory aware method is compared against two baselines.}
\label{tab:transition}
\resizebox{\columnwidth}{!}{%
\begin{tabular}{l|ccc|c}
\toprule
\textbf{Method} & \textbf{\shortstack{Geometric \\ Stability}} & \textbf{\shortstack{Smooth \\ Movement}} & \textbf{\shortstack{Visual \\ Pleasing}} & \textbf{\shortstack{Win \\ Rate}} \\
\midrule
Simple Concatenation & $7.0 \pm 0.9$ & $6.3 \pm 1.2$ & $5.6 \pm 1.0$ & 14\% \\
Blind Interpolator & $4.6 \pm 1.3$ & $5.9 \pm 1.2$ & $4.4 \pm 1.3$ & 7\% \\
w.o. adaptive windowing & $6.0 \pm 1.3$ & $6.4 \pm 1.0$ & $6.6 \pm 1.1$ & 15\% \\
\textbf{Trajectory Aware (Ours)} & \textbf{8.1} $\pm$ 1.0 & \textbf{7.6} $\pm$ 1.1 & \textbf{8.8} $\pm$ 1.1 & \textbf{64\%} \\
\bottomrule
\end{tabular}
}
\end{table}
\noindent
As shown in Table \ref{tab:transition}, our trajectory aware method decisively outperforms all baselines, confirming its necessity to produce continuous, cinematic motion. Also, for qualitative comparison, we bring trajectories of some samples before and after applying the adaptive transition in the supplementary \ref{sup: transition sample}.

\subsection{Discussion}
\label{sec:discussion}


Our findings in Section \ref{sec:exp_flf2v} (Table \ref{tab:video_gen_comparison}) are particularly revealing. While SOTA T2V models are highly competitive in single frame \textit{Realism Score} and \textit{Temporal Coherence}, they exhibit a clear failure in \textit{Camera Instruction Adherence}. This quantifies a critical trade off: current end to end models are optimized for visual and temporal plausibility, but they struggle to simultaneously manage scene content, composition, and specifically camera instruction.

Furthermore, this staged methodology offers advantages beyond the adherence metrics. The architecture inherently aligns with established, real world production pipelines in animation and cinematography \cite{hart2013art}. This alignment transforms the generation process from an end to end black box into an interpretable and controllable workflow. Our decoupled design naturally facilitates human in the loop intervention, and user can validate, refine, or manually edit the generated screenplay or the storyboard keyframes before committing to the computationally expensive video generation step. Each module's output serves as a distinct, verifiable artifact. This modularity not only makes the system understandable, but also allows for targeted validation and independent training or upgrading of components; such as swapping the T2I model or the FLF2V generator, without needing to retrain the entire system.




\section{conclusion}
We introduced CineLOG, a training free pipeline that addresses the poor cinematic control in end to end T2V models. We demonstrate that while SOTA models produce high fidelity visuals, they fail to adhere to explicit camera and screenplay directives. Our approach validates a decoupled framework which yields superior controllability without sacrificing visual quality. Our novel Adaptive Trajectory Guided Transition Module estimate spatio-temporal states at clip boundaries and construct a continuous path as a smooth control signal, ensuring coherent multi shot motion.
Finally, we release the CineLOG dataset: 5,000 balanced, high quality video clips annotated across 17 camera movements and 15 genres.

\clearpage
\setcounter{page}{1}
\maketitlesupplementary
\appendix

\section{Other Related Work}
\label{app: other related work}
Several other influential datasets provide signals that are related to camera control but are not directly suited for training cinematic camera controllers.

Efforts prioritizing domain breadth include Sekai \cite{li2025sekai}, with over 5,000 hours of FPV and drone footage with trajectories and rich metadata, and DropletVideo-10M \cite{zhang2025dropletvideo}, which contains 10 million clips with descriptive captions that often include camera actions, aimed at spatio-temporal consistency modeling. Human-centric corpora like HumanVid \cite{wang2024humanvid} use SLAM to extract camera paths from internet videos and pair them with synthetic avatars to train camera-aware human animation models.

Other datasets are tailored to different computer vision tasks. PointOdyssey \cite{zheng2023pointodyssey} is a long-form synthetic dataset with dense point correspondences designed for long-term tracking, not cinematography. Similarly, CamVid \cite{brostow2009semantic} provides labeled driving videos for semantic segmentation but features limited camera movement variety. Although these resources are valuable references for 6-DoF estimation and long-range tracking pipelines, they do not directly address the need for cinematically diverse and taxonomically labeled data essential for training versatile camera controllers.

\section{Failure T2V Models}
\label{app: failure models}

As stated in Section 3 of our main paper, we empirically observed that current end to end T2V models, regardless of their class, prioritize aesthetic plausibility over strict adherence to explicit cinematographic directives. This supplementary section provides qualitative evidence to support this claim, which is quantified in Table 4 of the main paper.

We present a visual comparison using a challenging screenplay generated by our pipeline's first stage (Section 3.1). This screenplay includes a detailed scene description, a specific genre, and a multi shot sequence with complex camera instructions (e.g., Shot 1: Truck Right, Shot 2: Dolly In). We provide this identical, complex prompt to two classes of SOTA T2V models and compare their output to our own.

\begin{itemize}
    \item \textbf{Class 1 (Open Source):} We evaluate \textbf{Wan 2.2} \cite{wan2025wan} and LTX model \cite{hacohen2024ltx} which are two of the best available SOTA models .
    \item \textbf{Class 2 (Premium Closed):} We evaluate \textbf{Hailuo 2.3 AI} \cite{HailuoAl2025} and \textbf{Veo 3.2} \cite{veo32025}, which are both released less than a month (when writing this paper).
\end{itemize}

\begin{figure*}
    \centering
    \includegraphics[width=\linewidth]{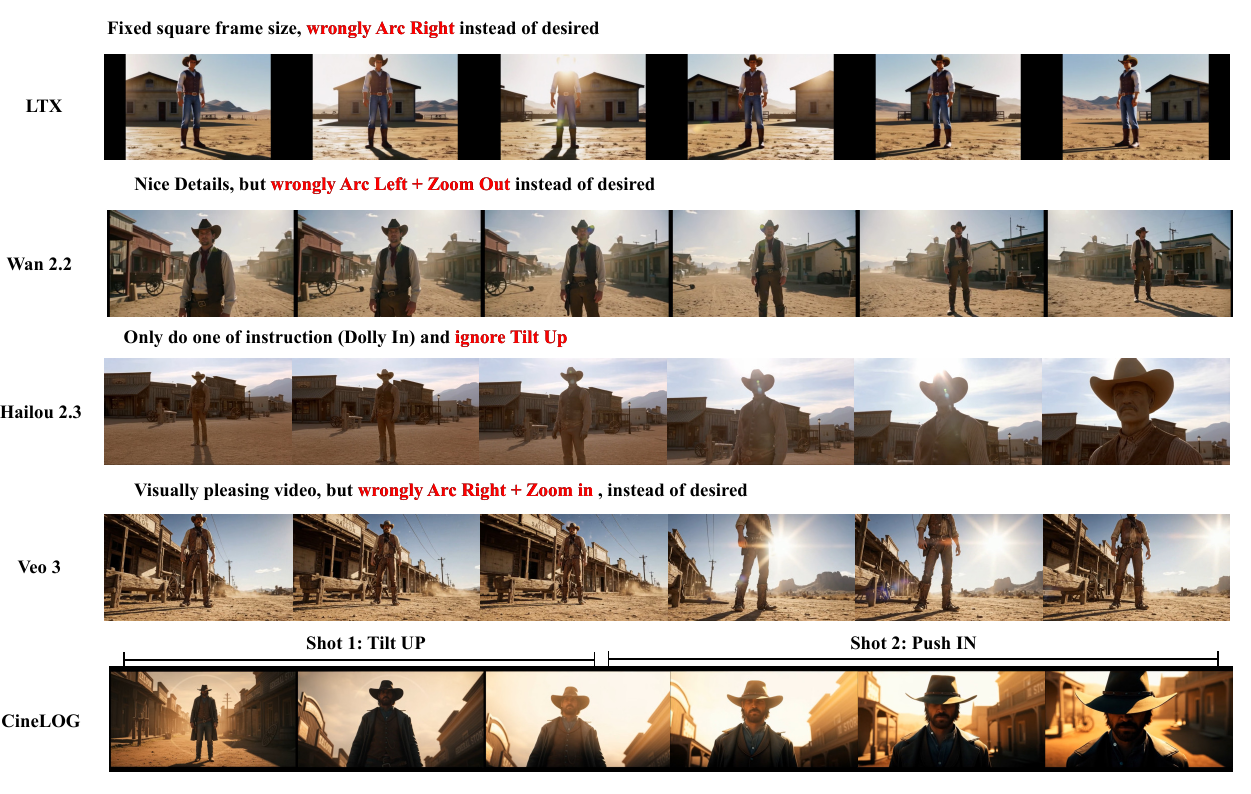}
\caption{Qualitative comparison of SOTA T2V models on Narration and Camera Adherence}
    \label{fig:failure_case}
\end{figure*}

Figure \ref{fig:failure_case} illustrates one prompt sample input to all models. While all models produce high fidelity, visually coherent clips, they consistently fail to execute the specified cinematic instruction.
The input prompt for all models is this:
\begin{description}
    \item[\textit{Narrative Description:}] 
    \textit{A tall cowboy stands statue like center on a dusty hot western street}
    
    \item[\textit{Shot Movements:}] 
    \textit{Shot 1: Tilt up movement; Cowboy remains centered, seen from a low upward tilt.}
     \textit{Shot 2: Push in; Camera dollies in to a half‑body, eye level framing of the cowboy.}
\end{description}

As seen in Figure \ref{fig:failure_case}, the T2V models often render the narration well, but almost all ignore the camera movement and instead default to a simple, subtle shot. Also, The models fail to adhere to the explicit Shot Count, and almost all multi shot instruction is collapsed into a single, continuous shot that vaguely blends the two concepts, demonstrating a clear failure in following precise screenplay structure.

In contrast, the CineLOG row in Figure \ref{fig:failure_case} demonstrates the success of our decoupled pipeline. By first generating precise keyframes and then interpolating between them, our method faithfully executes the specified camera trajectory and adheres to the multi shot structure, validating our central hypothesis.

\section{Screenplay Control Signal Details}
\label{app: screenplay details}
To ensure a diverse and balanced dataset, we explicitly defined and sampled a set of discrete control signals. These signals were used as conditional inputs for the Screenplay generation stage to avoid the mode collapse observed when tasking LLMs with random generation. The definitions and complete range of options for each signal are provided below.

\paragraph{Film Genre.} This signal defines the overall stylistic and narrative tone of the scene, influencing lighting, setting, and subject actions. We enforced a uniform balance across the following 13 distinct genres: \texttt{documentary}, \texttt{drama}, \texttt{action}, \texttt{comedy}, \texttt{horror}, \texttt{romance}, \texttt{fantasy}, \texttt{western}, \texttt{classic}, \texttt{animals}, \texttt{sports}, \texttt{science\ fiction}, and \texttt{war}.

\paragraph{Shot Count.} This variable determines the number of continuous shots, and thus the number of sequential keyframes, in a single video sequence. This allows for the creation of both simple, single action clips and more complex, multi part narratives. The value was selected from the discrete range: \texttt{1}, \texttt{2}, or \texttt{3} shots.

\paragraph{Camera Movement.} This parameter specifies the per shot camera motion. For multi shot sequences, a different movement was selected for each shot to ensure dynamic variety. Our taxonomy includes 17 well known cinematic movements: \texttt{static}, \texttt{pan left}, \texttt{pan right}, \texttt{tilt up}, \texttt{tilt down}, \texttt{dolly in}, \texttt{dolly out}, \texttt{truck left}, \texttt{truck right}, \texttt{pedestal up}, \texttt{pedestal down}, \texttt{zoom in}, \texttt{zoom out}, \texttt{crane up}, \texttt{crane down}, \texttt{arc left}, and \texttt{arc right}.

\paragraph{Main Subject Count.} This parameter controls the number of primary subjects or actors within the scene, which directly impacts the scene's composition and the complexity of the action. The values are defined as: \texttt{zero}, \texttt{single}, and \texttt{multiple}.

\paragraph{Scene Dynamicity.} This signal describes the motion state of the scene's content, independent of the camera's movement that has two type: \texttt{Static} scene involves subjects that are largely still. \texttt{Dynamic} scene involves significant subject motion.

\section{Evaluation Protocol}
\label{sup: evaluation protocol}
In this section we delve into details of human evaluations in our experiments.

\subsection{Storyboard Stage}

We evaluate all seven models across 100 screenplay samples using 30 human evaluators. For each sample, evaluators are presented with an input image, a target camera movement, and a narration describing the next scene. They then review the anonymized model outputs and rate them on a scale from 0 (Poor) to 10 (Perfect) according to three criteria:

\begin{enumerate}[label=(\roman*)]
\item \textbf{Camera Adherence:} Measures how accurately the model executes the specific camera instruction. 
\item \textbf{Scene Preservation:} Evaluates the visual consistency of the generation, specifically whether the subjects, objects, and background remain consistent with the input frame. 
\item \textbf{Narration Adherence:} Assesses the alignment between the generated image and the provided textual story description. 
\end{enumerate}

\subsection{FLF2V Stage Evaluation} 
In the evaluation of this stage, we recruited 30 expert evaluators to assess the outputs from all three methods. Participants were presented with the full screenplay and the corresponding three anonymized video outputs. We requested feedback in two distinct formats. First, evaluators provided scores on a scale from 0 (Poor) to 10 (Perfect) for the following qualitative metrics: \begin{enumerate}[label=(\roman*)] \item \textbf{Realism Score:} Evaluates the photorealism of the generated video, ensuring the content appears natural and of high visual fidelity. \item \textbf{Screenplay Adherence:} Measures the alignment between the visual content and the textual narrative description. \item \textbf{Camera Instruction Adherence:} Assesses the accuracy with which the generated video executes the specified camera movement commands. \item \textbf{Movement Naturalness:} Evaluates the fluidity of the scene and subject dynamics. \item \textbf{Geometric Stability:} Quantifies temporal consistency, specifically measuring the absence of flickering, texture warping, or geometric artifacts. \end{enumerate}

Second, evaluators assigned a binary label (Correct/1 or Incorrect/0) for each of the main control signals—Genre, Subject Count, Dynamicity, and Shot Count—to indicate whether the model successfully respected these specific constraints. These binary labels were aggregated to report the control accuracy for each model.

\section{Failure Automatic StoryBoard Evaluation}
\label{sec:supp_failure_metrics}

A natural starting point for assessing storyboard quality is to rely on automatic vision language metrics. Before designing our human evaluation protocol, we attempted to build a fully automated comparison pipeline for both Text to Image (T2I) models (for generating the initial keyframe) and Image to Image (I2I) models (for generating subsequent keyframes for multi shot clip). However, as discussed in Subsection \ref{sec:exp_storyboard}, these metrics proved fundamentally insufficient for evaluating the semantic correctness, visual continuity, and cinematographic precision required for generating our dataset. Below we describe the full experimental setup and present the quantitative results that motivated our shift toward human evaluation.

\subsection{T2I Benchmarking}

In the first experiment, we evaluated the capability of models to generate the initial shot based on the Screenplay's scene description. The results are presented in Table \ref{tab:T2I_automatic_comparison}.
 We inputted the scene prompt into various T2I models and evaluated the outputs using the following metrics:
\begin{itemize}
    \item \textbf{$CLIP_P$ Score:} Measures the semantic alignment between the generated image and the input text prompt.
    \item \textbf{Aesthetic Score:} Quantifies the perceived visual quality using the LAION aesthetic predictor \cite{schuhmann2022laion, schuhmann2022improved}.
    \item \textbf{Semantic Caption Similarity:} To assess how well the generated visual content reflects the narrative details, we employed a Vision Language Model (GPT-5 mini \cite{OpenAI2025GPT5}) to caption the generated images. We then computed the similarity between this generated caption and the ground truth input prompt using standard NLP metrics: \textbf{ROUGE-L}, \textbf{METEOR}, and \textbf{BERTScore}.
\end{itemize}

\begin{table*}[h]
\centering
\caption{Automatic evaluation results for T2I models.}
\label{tab:T2I_automatic_comparison}
\resizebox{\textwidth}{!}{%
\begin{tabular}{lccccc}
\toprule
\textbf{Model} & \textbf{$CLIP_P$} & \textbf{$Aesthetic$} & \textbf{$ROUGE-L$} & \textbf{$METEOR$} & \textbf{$BERTScore$} \\
\midrule
Bria Fibo \cite{gutflaish2025generating} & $32.31 \pm 0.20$ & $62.72 \pm 3.08$ & $26.98 \pm 0.18$ & $33.36 \pm 1.04$ & $\mathbf{88.77} \pm 0.01$ \\
Flux Schnell \cite{flux1schnell} & $33.95 \pm 0.08$ & $63.58 \pm 1.70$ & $23.64 \pm 0.12$ & $30.17 \pm 0.23$ & $88.12 \pm 0.02$ \\
Gemini Flash 2.5 \cite{GoogleGemini2.5FlashImage2025} & $\mathbf{34.13} \pm 0.24$ & $\mathbf{63.80} \pm 2.45$ & $26.86 \pm 0.22$ & $34.99 \pm 0.31$ & $88.58 \pm 0.01$ \\
Qwen Image \cite{wu2025qwen} & $33.81 \pm 0.22$ & $62.04 \pm 0.79$ & $29.80 \pm 0.81$ & $\mathbf{40.00} \pm 0.38$ & $88.31 \pm 0.02$ \\
Seedream 4 \cite{seedream2025seedream} & $33.04 \pm 0.19$ & $62.22 \pm 0.76$ & $\mathbf{33.52} \pm 0.46$ & $37.65 \pm 0.25$ & $88.69 \pm 0.03$ \\
\bottomrule
\end{tabular}%
}
\end{table*}

Despite covering several complementary metrics, the results provide no meaningful separation among models. All models exhibit similar prompt alignment, comparable aesthetic scores, and closely clustered caption similarity metrics. Crucially, none of these metrics correlate with failure cases in cinematic composition and camera aware framing, required for our storyboard generation.

\subsection{I2I Benchmarking}

In the second experiment, we assessed the performance of models in the sequential generation task, where applying desired edit and visual consistency is paramount. To ensure a fair comparison and mitigate bias from the initial generation, we first generated a starting frame using a randomly selected T2I model. This base image, along with the edit prompt (the description for the subsequent frame in our pipeline), was then passed to all evaluated I2I models.

In addition to the metrics used in the T2I experiment, we introduced \textbf{$CLIP_I$ Score}, which measures the cosine similarity between the CLIP embeddings of the reference image and the generated image. This metric serves as a proxy for checking visual consistency in our comparison. The results are summarized in Table \ref{tab:I2I_automatic_comparison}.


\begin{table*}[h]
\centering
\caption{Automatic evaluation results for I2I models.}
\label{tab:I2I_automatic_comparison}
\resizebox{\textwidth}{!}{%
\begin{tabular}{lcccccc}
\toprule
\textbf{Model} & \textbf{$CLIP_P$} & \textbf{$Aesthetic$} & \textbf{$ROUGE-L$} & \textbf{$METEOR$} & \textbf{$BERTScore$} & \textbf{$CLIP_I$} \\
\midrule
Bria Fibo \cite{gutflaish2025generating} & $32.49 \pm 0.03$ & $59.10 \pm 0.50$ & $25.47 \pm 1.10$ & $\mathbf{32.79} \pm 1.82$ & $88.24 \pm 0.02$ & $77.09 \pm 2.36$ \\
Flux Kontext Pro \cite{batifol2025flux} & $30.49 \pm 0.18$ & $61.50 \pm 1.39$ & $20.95 \pm 0.46$ & $23.37 \pm 0.28$ & $86.82 \pm 0.01$ & $\mathbf{88.03} \pm 0.90$ \\
Gemini Flash 2.5 \cite{GoogleGemini2.5FlashImage2025} & $\mathbf{32.53} \pm 0.09$ & $61.62 \pm 0.70$ & $25.59 \pm 0.74$ & $30.88 \pm 0.50$ & $87.51 \pm 0.02$ & $84.95 \pm 0.44$ \\
Qwen ImageEdit \cite{wu2025qwen} & $32.41 \pm 0.06$ & $\mathbf{62.30} \pm 1.64$ & $21.81 \pm 0.33$ & $25.93 \pm 0.18$ & $87.26 \pm 0.01$ & $83.66 \pm 1.21$ \\
SeedEdit 3.0 \cite{wang2025seededit} & $30.29 \pm 0.17$ & $60.32 \pm 0.93$ & $20.82 \pm 0.65$ & $26.06 \pm 0.86$ & $87.34 \pm 0.03$ & $81.74 \pm 1.89$ \\
Seedream 4 \cite{seedream2025seedream} & $31.98 \pm 0.08$ & $61.02 \pm 0.47$ & $\mathbf{27.57} \pm 0.60$ & $30.76 \pm 0.72$ & $\mathbf{88.29} \pm 0.01$ & $82.97 \pm 1.61$ \\
\bottomrule
\end{tabular}%
}
\end{table*}

Again, all models score similarly across semantic and aesthetic metrics. While $CLIP_I$ provides some signal on identity drift, it still fails to distinguish models in a way that correlates with Cinematography Adherence, Spatial Continuity, or Multishot Consistency, central properties required for our pipeline.

As demonstrated in Tables \ref{tab:T2I_automatic_comparison} and \ref{tab:I2I_automatic_comparison}, the quantitative results exhibit a high degree of saturation. The performance gaps between models are marginal, with scores clustered tightly together (e.g., BERTScore differences are negligible across all models). Furthermore, these metrics primarily measure general image quality and semantic relatedness but fail to account for the specific spatial and structural constraints required by our cinematic taxonomy (e.g., distinguishing a "Dolly In" from a "Zoom In").

Given that the automatic metrics could not provide a clear separation between models or reliably indicate adherence to complex camera instructions, we concluded that they were inadequate for selecting the optimal generator for our pipeline. Consequently, we transitioned to the rigorous human evaluation protocol to ensure the CineLOG dataset achieves the highest possible professional standard.

\section{Qualitative Validation of Trajectory Guidance}
\label{sup: transition sample}

\begin{figure*}[t]
\centering
\includegraphics[width=\linewidth]{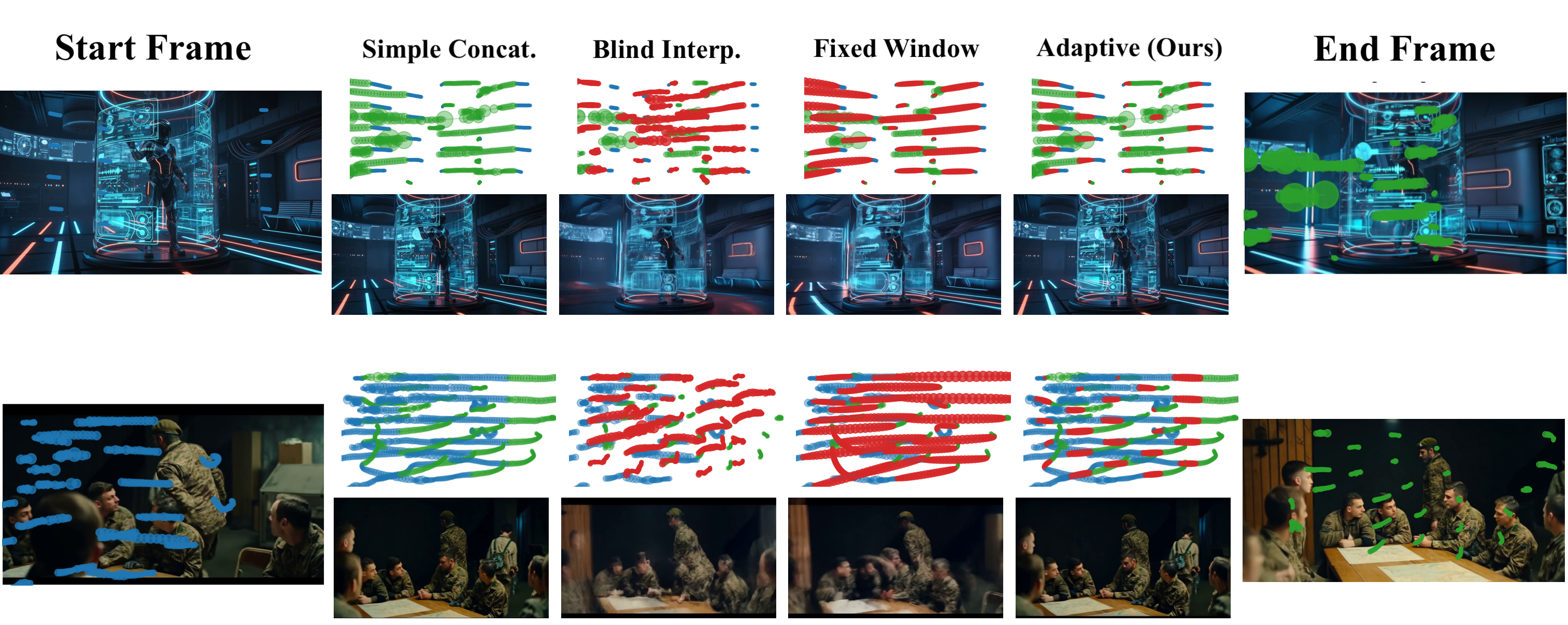}
\caption{Qualitative comparison of transition strategies across two samples. For each method, the top row illustrates the motion trajectory (dot size indicates velocity magnitude), and the bottom row displays the generated intermediate frame. The blue color shows trajectory in start frame, blue trajectory in end frame and red denotes the interpolated trajectory. Note that while Simple Concatenation maintains high image sharpness, it suffers from severe kinematic discontinuities (trajectory hiccups). Conversely, baseline interpolation methods result in blurry, noisy frames due to distribution shifts. Our Adaptive Strategy creates the optimal balance, preserving the perceptual fidelity of the frames while ensuring a smooth, stable trajectory.}
\label{fig:transition_viz}
\end{figure*}

To qualitatively validate the impact of our Trajectory Guided Transition Module, we present a visualization of the transition boundary in Figure \ref{fig:transition_viz}. The figure illustrates two distinct samples. For each sample, we provide a dual row visualization: the top row plots the trajectory line, where the size of each point corresponds to the velocity magnitude, while the bottom row displays the generated intermediate frame from the transition sequence.

Visual inspection reveals critical disparities between spatial image quality and temporal motion stability across the different methods. As expected, Simple Concatenation exhibits high perceptual fidelity and sharpness, as the individual frames are generated directly by T2I models without interference; however, the trajectory visualization reveals an immediate kinematic fracture where the camera position resets, resulting in a lack of temporal coherence. Conversely, while Baseline Interpolation (Blind \& Fixed-Window) attempts to bridge this temporal gap, it significantly degrades visual quality. The intermediate frames suffer from noticeable blurring and high-frequency noise, largely attributed to pixel level distribution shifts between the start and end frames which destabilize the underlying flow estimation. In strong contrast, our Adaptive Trajectory Guided Strategy demonstrates the most robust performance by effectively balancing spatial and temporal quality. Our method preserves the sharp, photorealistic quality of the frames, similar to Simple Concatenation, while simultaneously eliminating trajectory fractures. By respecting boundary velocities, it constructs a smooth, continuous path, yielding a seamless cinematic progression free from both mechanical speed ramps and visual blurring.

\newpage

{
    \small
    \bibliographystyle{ieeenat_fullname}
    \bibliography{main}
}


\end{document}